\documentclass[runningheads,a4paper]{llncs}

\usepackage[utf8]{inputenc}

\usepackage[english]{babel}
\usepackage[backend=bibtex,style=ieee,sorting=ynt]{biblatex}
\usepackage{amsmath}
\usepackage{amssymb}
\usepackage{graphicx}

\usepackage[english]{babel}
\usepackage[T1]{fontenc}

\usepackage{url}
\urldef{\mailsa}\path|{alfred.hofmann, ursula.barth, ingrid.haas, frank.holzwarth,|
\urldef{\mailsb}\path|anna.kramer, leonie.kunz, christine.reiss, nicole.sator,|
\urldef{\mailsc}\path|erika.siebert-cole, peter.strasser, lncs}@springer.com|

\begin{document}

\mainmatter 

\title{Deep Reinforcement Learning Radio Control and Signal Detection with KeRLym, a Gym RL Agent}

\titlerunning{RL Radio Control \& KeRLym}

\author{Timothy J O'Shea and T. Charles Clancy}

\institute{Virginia Polytechnic Institute and State University}

\authorrunning{T. O'Shea, T. C. Clancy}

\maketitle

\begin{abstract}
This paper presents research in progress investigating the viability and adaptation of reinforcement learning using deep neural network based function approximation for the task of radio control and signal detection in the wireless domain.  We demonstrate a successful initial method for radio control which allows naive learning of search without the need for expert features, heuristics, or search strategies.  We also introduce Kerlym, an open Keras based reinforcement learning agent collection for OpenAI's Gym.
\end{abstract}

\medskip

\section*{Introduction}

Radios are ubiquitous in modern society.  Between cellular devices, wearable devices, computing devices, medical devices, and other devices we carry and operate regularly through each day, radio frequency communications have become the most pervasive and convenient way we communicate information in our daily lives.  Unfortunately, our spectrum resources are limited and our data needs are growing in a seemingly unbounded manner.  Despite this wireless spectrum crunch, our methods for allocation, adaptation and optimization of spectrum use remains very much in the dark ages today.  Spectrum is still allocated in a static fashion, and devices are oblivious and unaware of the use of or the availability of resources in their direct vicinity.  The field of cognitive radio and dynamic spectrum access have attempted to address this through the introduction of expert systems which attempt to perform spectrum sensing and some degree of characterization of their environment, but their impact has been heavily limited by their inability to generalize to new regions, protocols, emitters, and radio propagation environments.

Generalized policy learning has and continues to be an open challenge in CS and AI for many years, however in recent years advances in reinforcement learning have made massive strides towards the advancement of this field.  Recent Work by Minh \cite{mnih2013playing}, Silver \cite{nature-go}, Sutton \cite{sutten_textbook}, and others has begun to demonstrate the ability to learn exceedingly complex and varied tasks using deep neural network based policy function approximation to implement Q-Learning.  

To address this problem of learning to rapidly understand the surrounding radio environment, we introduce a radio signal search environment for the recently released Gym RL framework from OpenAI in which to begin evaluating and scoring different approaches.

We also implement a general purpose open source Deep Neural network based Q-Learning function approximation learner for Gym using Keras primitives to learn a policy for rapidly exploring this environment through its set of discrete actions and observations \cite{keras}.  

\section{Reinforcement Learning Policy}

We introduce KeRLym \cite{kerlym}, an open source deep reinforcement learning agent collection written in python using Keras to implement GPU optimized deep neural networks on top of Theano \cite{theano} and TensorFlow \cite{tensorflow}.  OpenAI recently released Gym \cite{duan2016benchmarking}, a collection of reinforcement learning benchmark environments, an API to easily use them, and a web based high-score board for algorithm comparison.   We leverage this API in our reinforcement learner to provide a standard agent interface and to rapidly provide a wide range of tasks we can test its performance and tuning against.

\subsection{Policy Learning}

Since Google Deepmind's Nature paper on Deep-Q Networks \cite{mnih2013playing}, there has been a surge of interest in the capabilities of reinforcement learning use deep neural network policy network approximation.   This is an exciting and growing area with much potential improvement for learning algorithms yet to come.   For the scope of this work we implement a parametric version of the Deep Q-Learning algorithm along with a Double Q-Learning \cite{van2015deep} implementation in the KeRLym toolbox.   We implement a variety of function approximation networks which can be used inside them including dense fully connected networks, convolutional networks similar to those used in the Atari paper, and recurrent networks leveraging LSTM which may improve sequence learning in POMDPs as discussed in \cite{hausknecht2015deep}.

The approaches in are similar for both algorithms, a value function $Q(s,a;\theta)$ is updated using a form of stochastic gradient descnt, SGD, in the form of:

\begin{equation}
    \theta_{t+1} = \theta_t + \alpha \left( Y_t, - Q\left( S_t, A_t; \theta_t \right) \right) \nabla_\theta Q(S_t, A_t; \theta_t)
\end{equation}

However, in single Q-learning we directly compute $Y_t$ in a greedy manner using our latest $\theta$ as:

\begin{equation}
    Y_t = R_{t+1} + \gamma \underset{a}{ \mathrm{max} } Q\left(S_{t+1}, a; \theta_t\right) 
\end{equation}

Whereas in double Q-learning we maintain two sets of weights $\theta$ and $\theta'$ which we alternate between using for decision making and greedy policy update purposes:

\begin{equation}
    Y_t = R_{t+1} + \gamma Q \left( S_{t+1}, \underset{a}{ \mathrm{argmax} } Q\left(S_{t+1}, a; \theta_t\right) ; \theta' \right) 
\end{equation}

This helps to reduce overestimation value bias and imrpoves policy learning rate and stability for many tasks.

We implement $\epsilon$-greedy learning with a default constant value of 0.1, to choose the greedy policy 90\% of the time, simply to avoid the tuning required with $epsilon$ decay schedules for stability of comparison of this work.

We also implement experience replay, keeping around a memory of 1,000,000 previous actions to draw training samples from in addition to the new experience gained each time-step.  We use a learning rate of 0.001 in a Keras Adam \cite{adam} solver, and a discount rate of $\gamma = 0.99$ in our experiments.

Within the KeRLym toolbox we hope to extend the number of solvers available 

\subsection{Deep-Q Network Implementation}

Our Q function $Q(s,a,\theta)$ is a Deep Neural Network with random initial parameters $\theta$ implemented in the Keras framework on top of Theano, running on an Nvidia Titan X.  We zero the output regression layer weights to reduce initial error in value function output.  We start with a similar architecture to the convolutional network used by Mnih et al in \cite{mnih2013playing}, but make changes which show improvement in our domain and account for the input information form.

Since we are passing both scalar stored variables containing sensor information, and contiguous frequency domain values into the value function as the current state, we treat each input configuration value as an independent discrete input with fully connected logic, while we reduce the parameter space and allow frequency domain filters to form and be used shift-invariantly on the power spectrum by using a set of convolutional layers, similar to our approach on raw time-domain samples in \cite{convmodrec}.

Ultimately we concatenate the activations from both of these paths into dense fully connected layers to perform the output regression task for output action-value estimates.

\begin{figure}[ht!]
    \centering
    \includegraphics[width=1.0\textwidth]{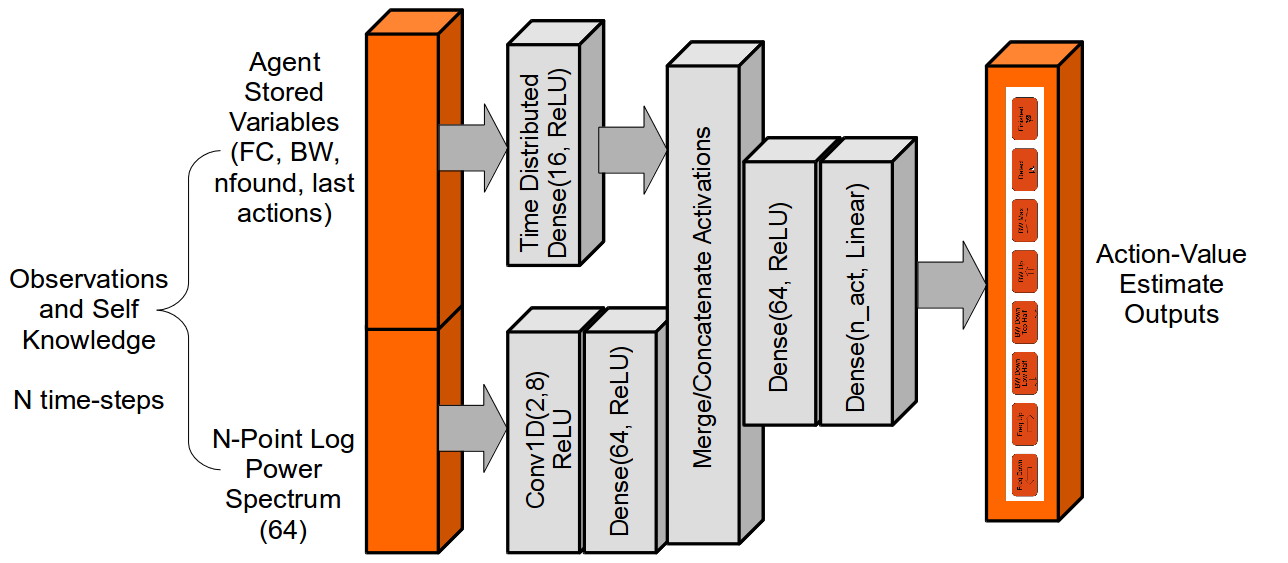}
    \caption{Action-Value Network Architecture}
    \label{fig:net}
\end{figure}

\section{Radio Search Environment}
\subsection{Environment Overview}

Typical electronic devices such as cellular phones contain at this point highly flexible Radio Frequency Integrated Circuits (RFIC) which allow the frequency tuning and digitization of relatively large arbitrary bands of interest.  Typically they are programmed in an exceedingly simple way by a carrier to brute force through a small list of carrier-assigned channels and bandwidth, however they are in fact capable of tuning to relatively arbitrary center frequencies between 100 MHz and 6GHz and providing often powers of two decimations of a 10-20 MHz wide bandwidth.   

Instead of brute force search for signals on several carrier centric bands, we propose instead to allow machine learning to derive a general search policy to identify signals providing useful connectivity while optimizing for minimal search time, battery consumption and power usage possible.  To do this we boil the search task down into a relatively small set of possible discrete actions which may be taken towards the end-goal.   

\begin{figure}[h]
    \centering
    \includegraphics[width=0.75\textwidth]{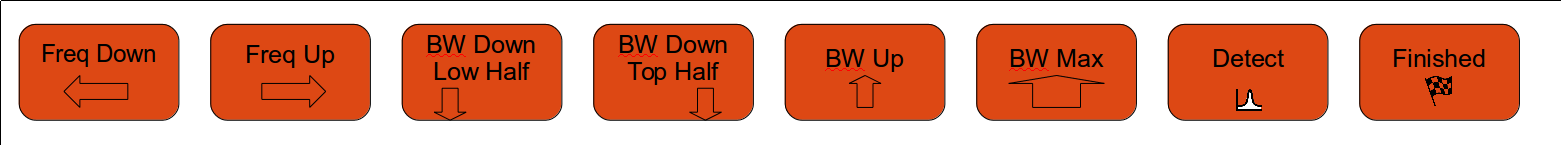}
    \caption{Initial Radio Discrete Action Set}
    \label{fig:actionset}
\end{figure}

\subsection{Environment Implementation}

We begin by building an environment for the Gym Reinforcement Learning environment to attempt to mirror our problem statement, and a reasonable set of assumptions for what a real system could do and sense, but at a relatively small scale of complexity for initial work.

We simulate a single radio receiver sampling at a bandwidth of 20 MHz, which can be decimated and re-tuned using the set of discrete actions in \ref{fig:actionset}.  The discrete actions we allow are as follows, where we refer to the variables: center frequency (fc), bandwidth (bw), maximum bandwidth (bwmax), minimum bandwidth (bwmin), maximum center frequency (fcmax), and minimum center frequency (fcmin).

\begin{itemize}
    \item Freq Down: $fc -= max(bw/2, fc_min)$
    \item Freq Up: $fc += min(bw/2, fc_max)$
    \item BW Down Left: $bw = max(bw/2, bw_min); fc -= bw/2$
    \item BW Down Left: $bw = max(bw/2, bw_min); fc += bw/2$
    \item BW Max: $bw = bw_max$
    \item Detect: Assert that a signal is in the current window.
    \item Finished: Assert that all signals in band have been detected.
\end{itemize}

\begin{figure}[ht!]
    \centering
    \includegraphics[width=0.75\textwidth]{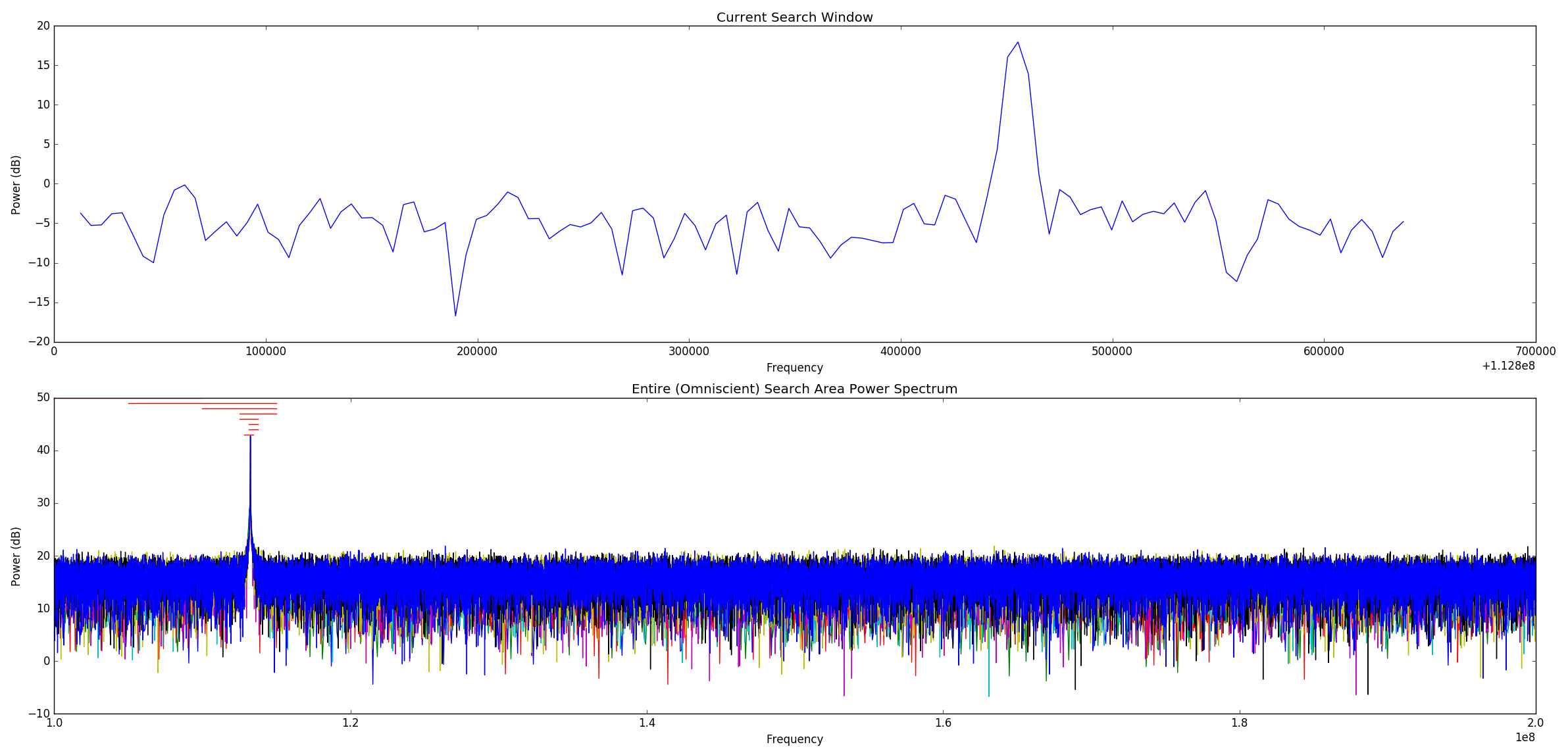}
    \caption{Environment Search Scenario}
    \label{fig:search}
\end{figure}

The environment chooses a random frequency within the band of interest (100MHz to 200MHz in this work) to place a single sinusoidal tone.  For each agent observation, it returns a small band-limited window into the environment tuned to the chosen center frequency and bandwidth.  The Detect action asserts that there is a signal within the current band either correctly or falsely, Finish assets that we have correctly found the signal and our search path is complete, and bandwidth and frequency actions change our receiver configuration according to the list above.   A single optimal path to a solution through the environment might look something like shown in figure \ref{fig:search}.  In this case each look window for a time-step is represented by a red bar above the wideband power spectrum plot.

\section{Training Considerations}

There are numerous ways to define penalties and rewards for this search process within the environment which pose a number of different considerations for the training process, we propose 3 potential rewards schemes below.

\begin{table}[!htb]
    \centering

 \begin{tabular}{||c c c c||} 
 \hline
 Scheme & A & B & C \\ [0.5ex] 
 \hline\hline
 Detect(True)   & 1 & 1 & 0 \\ 
 \hline
 Detect(False)  & 0 & -1 & 0 \\
 \hline
 BW-(True)      & 1 & 1 & 0 \\
 \hline
 BW-(False)     & 0 & 0 & 0 \\
 \hline
 Finish(True)   & 1 & 1 & nfound*depth \\ [1ex] 
 \hline
 Finish(False)  & 0 & -1 & 0 \\ [1ex] 
 \hline
\end{tabular}\\
\caption{Environment Reward and Penalty Schemes}

\end{table}

Oour agent's goal at run-time is to detect the signal present somewhere in the band and localize the signal using BW-L and BW-R actions to zoom in on it, these rewards and penalties are designed to reflect that.   
Scheme A results in perhaps the fastest training rate and simples approach towards directly rewarding good actions, Scheme B provides a strong disincentive for false positive actions, but slows down learning, and Scheme C provides a simple final score which requires a more delayed-reward style of learning.

\section{Conclusions and Future Work}

\begin{figure}[h]
    \centering
    \includegraphics[width=0.75\textwidth]{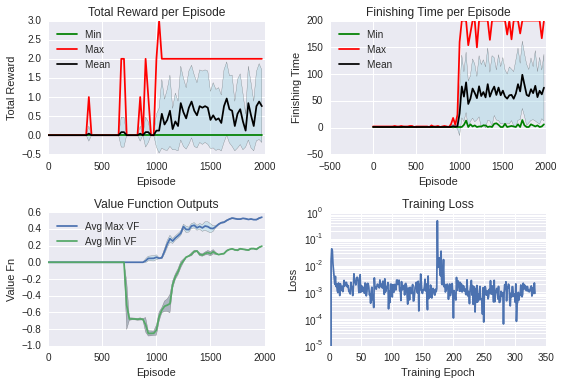}
    \caption{Plots During Network Training}
    \label{fig:plots}
\end{figure}

We can plot a number of statistics during the training process which give us insight into how the training is going.   Shown in figure \ref{fig:plots} we have the training statistics under Scheme B with early exiting (on Finish(False)).  From this graph it is clear that we are learning a relatively clear separation between good and bad action values, as can be seen from the separation in the 3rd plot, and that our reward is growing and our finishing time is growing long enough to succeed some of the time.

'
In future work we hope to provide a more comprehensive trade between the trade offs described above, learn a policy which performs at a more satisfying reward level, and and compare the impact of reward/penalty schemes on traditional receiver operating characteristics, ROC, curves for performance.  We are excited about the potential in this area and positive this approach will be fruitful.

\noindent

\printbibliography


\nocite{*} 

\end{document}